\documentclass[letterpaper, 10 pt, conference]{ieeeconf}  % Comment this line out
\IEEEoverridecommandlockouts                              % This command is only
\usepackage[utf8]{inputenc}
\usepackage{graphicx} % for pdf, bitmapped graphics files
\usepackage{amsmath} % assumes amsmath package installed
\usepackage{multirow}
\usepackage{graphicx}
\usepackage{setspace}
\usepackage{tikz}
\usepackage{float}
\usepackage{amsmath}
\usepackage{algorithm}
\usepackage{listings}
\usepackage{multirow} 
\usepackage{rotating}
\usepackage{color, colortbl}
\usepackage{pdfpages}
\usepackage{algorithmicx,algpseudocode}
\usepackage{relsize}
\usepackage{subcaption}

\usepackage{url} %Supports url commands

\algnewcommand\algorithmicswitch{\textbf{switch}}
\algnewcommand\algorithmiccase{\textbf{case}}
\algnewcommand\algorithmicassert{\texttt{assert}}
\algnewcommand\Assert[1]{\State \algorithmicassert(#1)}%
\algdef{SE}[SWITCH]{Switch}{EndSwitch}[1]{\algorithmicswitch\ #1\ \algorithmicdo}{\algorithmicend\ \algorithmicswitch}%
\algdef{SE}[CASE]{Case}{EndCase}[1]{\algorithmiccase\ #1}{\algorithmicend\ \algorithmiccase}%
\algtext*{EndSwitch}%
\algtext*{EndCase}%

\title{\LARGE \bf
Intelligent flat-and-textureless object manipulation in Service Robots
}

\author{
Abel Pacheco-Ortega$^{1}$, Hugo Estrada$^{1}$, Edgar Vázquez$^{1}$, Reynaldo Martell$^{1}$, Jesús Hernández$^{1}$, \\Julio Cruz$^{1}$, Edgar Silva$^{1}$, Jesus Savage$^{1,*}$ and Luis Contreras$^{2}$
	\thanks{The authors are with the $^{1}$Faculty of Engineering, National Autonomous University of Mexico and the $^{2}$Advance Intelligence and Robotics Research Center at Tamagawa University, Japan.}%
    \thanks{This work was supported by PAPIIT-DGAPA UNAM under Grant IG-100818}
	\thanks{$^{*}$savage at servidor.unam.mx }
}

\begin{document}

\maketitle
\thispagestyle{empty}
\pagestyle{empty}

%%%%%%%%%%%%%%%%%%%%%%%%%%%%%%%%%%%%%%%%%%%%%%%%%%%%%%%%%%%%%%%%%%%%%%%%%%%%%%%%
\begin{abstract}
This work introduces our approach to the flat and textureless object grasping problem. In particular, we address the tableware and cutlery manipulation problem where a service robot has to clean up a table. Our solution integrates colour and 2D and 3D geometry information to describe objects, and this information is given to the robot action planner to find the best grasping trajectory depending on the object class. Furthermore, we use visual feedback as a verification step to determine if the grasping process has successfully occurred. We evaluate our approach in both an open and a standard service robot platform following the RoboCup@Home international tournament regulations.
\end{abstract}

%%%%%%%%%%%%%%%%%%%%%%%%%%%%%%%%%%%%%%%%%%%%%%%%%%%%%%%%%%%%%%%%%%%%%%%%%%%%%%%%
\section{INTRODUCTION}

One way to evaluate the state-of-the-art on object manipulation is through standard setups in international competitions such as Robocup@Home \cite{wisspeintner:2009}, whose purpose is the development of Domestic Service Robots, with a high relevance in the future of domestic and assistance applications. Since these agents move and interact in non-structured spaces and objects, the range of strategies, algorithms, and approaches that each challenge implies is considerable.

Grasping objects from a table is a common task for Service Robots. Robots mounted with active sensors such as RGB-D cameras are able to detect the dominant plane, cluster point clouds on the plane and analyze them in order to detect the objects on it and determine the best strategy to manipulate the objects as in \cite{Rao2010},\cite{Holz2011} and \cite{Neves2016}. However, when the objects lack texture and or volume (i.e. flat objects), the manipulation task becomes more challenging.

%RoboCup is an international project whose purpose is to promote development in fields such as artificial intelligence and robotics through competencies where challenges are established in which universities and institutions from all over the world participate and compete.

\begin{figure}[ht]
	\centering
	\begin{subfigure}{.225\textwidth}
  		\centering
  		\includegraphics[width=1\textwidth]{{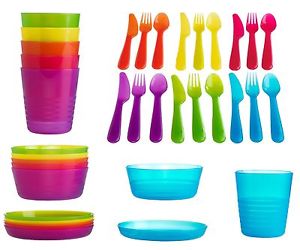}}
        \caption{}
  		\label{fig:rgbview}
	\end{subfigure}
	\begin{subfigure}{.225\textwidth}
  		\centering
  		\includegraphics[width=1\textwidth]{{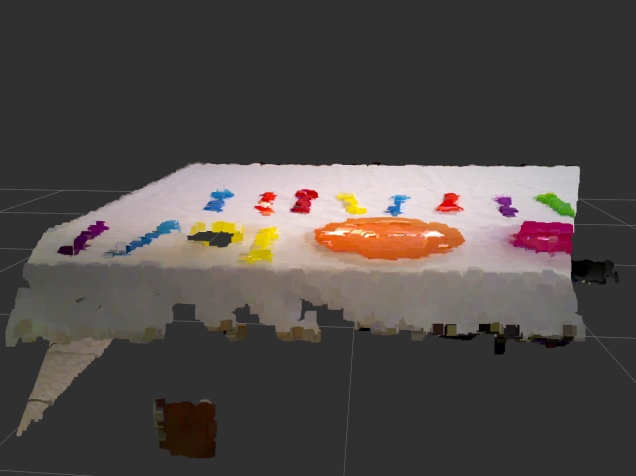}}
        \caption{}
  		\label{fig:depthview}
	\end{subfigure}
	\caption{a) Tableware and cutlery objects used in the RoboCup@Home's P\&G Challenge and b) table setup 3D view from an RGBD camera. It can be observed that the reconstructed scene presents some errors due to sensor noise, varying illumination conditions and high object reflectance.}
	\label{fig:rgbdview}
\end{figure}

In the 2018 edition of Robocup@Home, one of the tests was the Procter \& Gamble (TM) Dishwasher Challenge (PGDC) in which they have to remove all objects from a table and place them into a dishwasher \cite{rulebook_2018}. The set of objects that were used during the test are made of plastic, without any visual pattern on them but colours with high contrast, as shown in Figure \ref{fig:rgbview}. Besides, the objects' surface is very reflective provoking a significant variation in colour. In consequence, the information sensed by the RGB-D camera results noisy and the objects seem to belong to the table, as shown in Figure \ref{fig:depthview}.

In this work, we present our strategy to successfully manipulate the objects in the dishwasher challenge. In the next sections, we will detail the techniques used to detect, recognize and manipulate tableware and cutlery objects.

\section{Object detection and manipulation}

Typical object recognition strategies consist on object segmentation by plane extraction and point cloud features clustering  (\cite{Richtsfeld_2012}, \cite{uckermann_2012}, \cite{Ying_2017}, \cite{scstein_2014}) or object detection in the whole image by feature matching (\cite{Ikizler_2010}, \cite{Alhwarin_2008}, \cite{Lowe2004},\cite{Calonder_2010}). %\textbf{(FIX: Add some references)}
However, in situations where the objects are flat, plane extraction is no possible; similarly, when textureless objects are present, feature detection and matching become unavailable and other approaches need to be explored.

%Jesus Coyotzi: 
Another popular approach to solve this problem is the use of Deep Neural Networks, in particular the YOLO \cite{yolov3} architecture provides state of the art performance on the object detection task. Our approach does not rely on neural networks and thus we do not rely on specialized hardware such as GPU's or online servers. Also, we need much less training examples compared to YOLO, which needs a couple hundreds of images per object; in the same sense, YOLO requires labelled examples from different poses and in different backgrounds in order to generalize correctly, our approach uses just a couple images from the same pose in the same background. 

In this section, we present our solution to the flat and textureless object detection and recognition problem. We first use colour features in the detection process and then extract the geometry properties using the point cloud to describe them. This information also results useful in the manipulation process.

%\subsection{Training}

\begin{figure}[h]
\begin{center}
\resizebox{0.5\textwidth}{!}{
		$\begin{array}{c}
		\includegraphics[width=50mm]{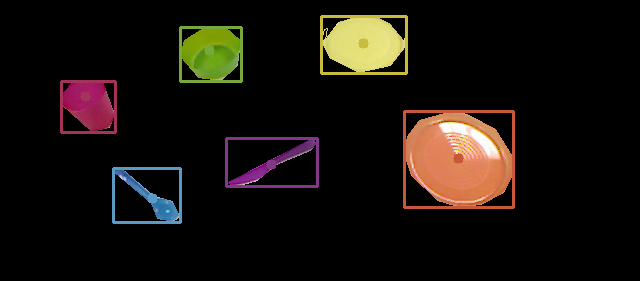} 
		\end{array}$}
	\end{center}
	\caption[]{Typical output after object detection using colour features.}
	\label{fig:Segmentation}
\end{figure}

For background subtraction, we find the range of values in each channel on three different colour spaces (RGB, HSV and HLS) for all objects and a raw segmentation mask is obtained where morphological operations of closing and dilation are applied to close the gaps where segmentation was wrong%, as shown in Figure \ref{fig:objectseg}.
To deal with cases where objects present significant illumination changes, a convex hull operation is applied as the final step in the segmentation process. Finally, we obtain an RGB mask and a 2D bounding box per object, as illustrated in Figure \ref{fig:Segmentation}.

%\begin{figure}[ht]
%	\centering
%	\begin{subfigure}{.225\textwidth}
% 		\centering 
%  		\includegraphics[width=1\textwidth]{{images/training1.jpg}}
%  		\caption{}
%  		\label{fig:colorseg}
%	\end{subfigure}
%	\begin{subfigure}{.225\textwidth}
%  		\centering
%  		\includegraphics[width=1\textwidth]{{images/training2.jpg}}
%  		\caption{}
%  		\label{fig:morphoseg}
%	\end{subfigure}
%	\caption{Arrangement used for training, a) different positions and intensities of light are used, b) three colour spaces are tested to determine which best defines the space occupied by the objects of the same colour.
    %\textbf{(FIX: Extend the caption)}.
%    }
%	\label{fig:objectseg}
%\end{figure}

%change this part: Through the area occupied for every object, the robot is capable of determining what kind of object is presented: dish, bowl, glass and cutlery.

%Once objects have been detected by colour and extracted certain properties such as the eigenvectors, eigenvalues, area of the object, number of points and the bounding box, we can categorize the objects into 4 different classes: glasses, plates, bowls and cutleries.%

We then categorise objects into four different classes, namely glass, plate, bowl, and cutlery, by normalising the image with respect to the average depth and then extracting the area in the RGB image and the number of points and the eigenvectors from the point cloud, as will be described next.

We consider the differences in the area among the objects -- for example, the glass-type object, despite having greater height, its area is not very large due to its transparency --, that can provide an initial hint on the object class. In a training step, from a set of segmented images, a valid area range per class is obtained to determine if it is a given category in case the object's area is within a valid range. %\textbf{FIX: Do we use area, point count, and geometry criteria all together or we use point count and geometry criteria and compare with the area criterion?, We use all together (area, geometry criteria and point count).}

However, due to changes in illumination, the visible area may change and therefore we also consider the 3D point count per object. From the same point of view, the dimensions of the cutlery and glass differ greatly from that of the plate or bowl, hence, we set a valid threshold among those classes, so if an object is within this range it is said that it can be in the cutlery and glass sub-category, and in the plate and bowl sub-category otherwise.

%To classify the object as cutlery or glass we observed that the eigenvalues of one of the axes of the cutlery are very small in comparison with the glasses, this due to the elongated geometry of the cutlery. With this feature we can differentiate between both objects.

%We will apply a third criterion to determine if it is a dish or a bowl. The third criterion is a principal component analysis (PCA) of the point cloud of the object, this consisted in using the eigenvalues. To classify only two axes are taken since a component of the objects has no height, the method consisted of obtaining the smallest principal component, it is worth noting that we compared the relationship between the two principal components of the bowls and plates, and having the same circular shape this relationship seemed a lot, so it did not help us in the categorization; however, the eigenvalues of both objects were compared, resulting in that the dishes have larger dimensions, the eigenvalues will be much larger than those of a bowl, so only comparing the smaller main component with a threshold, categorizes between a bowl and a plate.

\begin{figure}[h]
\begin{center}
\resizebox{0.50\textwidth}{!}{
		$\begin{array}{c}
		\includegraphics[width=50mm]{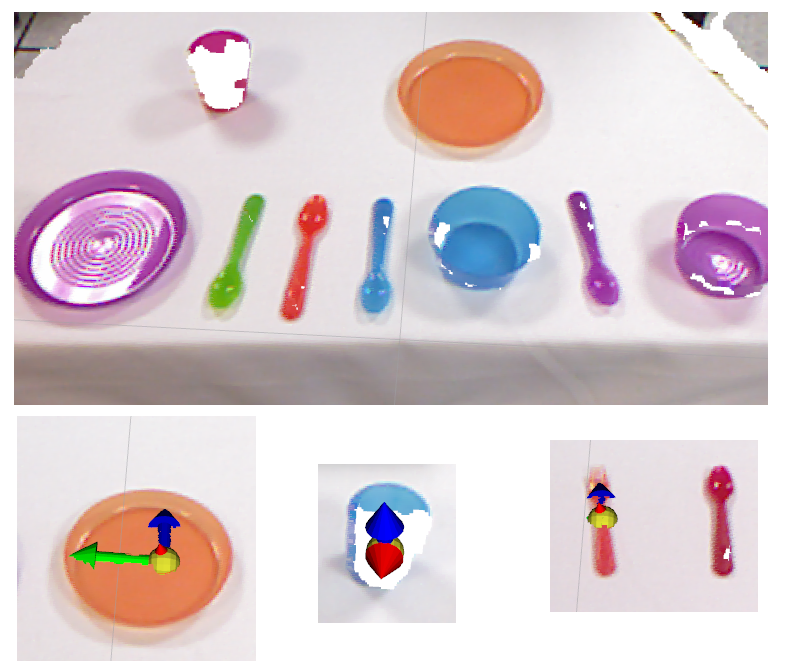} 
		\end{array}$}
	\end{center}
	\caption[]{ Eigenvectors for a group of common cutlery used in P\&G's test. Blue colour shows the eigenvector with bigger magnitude. Eigenvectors were obtained with PCA.}
	\label{fig:execution}
\end{figure}

\begin{figure}[h]
\begin{center}
\resizebox{0.50\textwidth}{!}{
		$\begin{array}{c}
		\includegraphics[width=50mm]{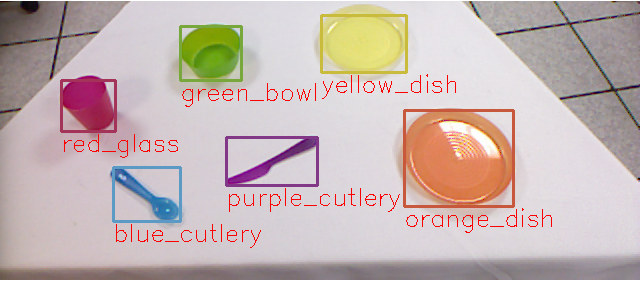} 
		\end{array}$}
	\end{center}
	\caption[]{Classification result using 3D information to the setup given in Figure \ref{fig:Segmentation}.}
	\label{fig:Classification}
\end{figure}

To correctly classify objects in the sub-categories, we consider their geometry properties as follows. From an object's point cloud of 3D features, we extract their eigenvector and eigenvalues using principal component analysis (PCA) and align them to the dominant plane coordinate system (plane XY and normal Z), as can be seen in Figure \ref{fig:execution}. For the cutlery and glass sub-category, we assign the cutlery class to any object with a small height and a small ratio of the shortest XY axis to the largest one, due to the flat and elongated shape, and glass class otherwise. In the case of the dish and bowl sub-category, we only consider the height component due to the similarity on the shape in the XY plane between a plate (flat, small Z) and bowl (deep, large Z). An example of the classification result is shown in Figure \ref{fig:Classification}.

%\textbf{TODO - Add PCA images and references.}\\

%\textbf{TODO - Report classification results.}\\
After testing 1500 times the algorithm under different conditions and using different object configurations we obtained the results presented in the following confusion matrix, where a high recognition rate is observed.\\

\setlength\unitlength{1.2cm}
\begin{center}
\begin{picture}(5,5) \label{matrix} 
\multiput(0.1,0.1)(0,1){5}{\line(1,0){4}}
\multiput(0.1,0.1)(1,0){5}{\line(0,1){4}}
\put(0.5,0.5){2\%}
\put(1.5,0.5){1\%}
\put(2.5,0.5){2\%}
\put(3.5, 0.5){95\%}

\put(0.5,1.5){3\%}
\put(1.5,1.5){5\%}
\put(2.5,1.5){91\%}
\put(3.5,1.5){1\%}

\put(0.5,2.5){2\%}
\put(1.5,2.5){90\%}
\put(2.5,2.5){6\%}
\put(3.5,2.5){2\%}

\put(0.5,3.5){93\%}
\put(1.5,3.5){2\%}
\put(2.5,3.5){3\%}
\put(3.5,3.5){2\%}

\put(-0.6,3.5){Glass}
\put(-0.6,2.5){Dish}
\put(-0.6,1.5){Bowl}
\put(-0.9,0.5){Cutlery}

\put(0.2,4.2){Glass}
\put(1.3,4.2){Dish}
\put(2.3,4.2){Bowl}
\put(3.2,4.2){Cutlery}

\put (-1.2,1.2){\rotatebox{90}{Actual Class}}
\put(1.2,5){Predicted Class}
\end{picture}
\end{center}

%\textbf{TODO - We can compare here with YOLO.}\\

\section{Interaction with real world}

%\subsection{Messages to other modules in the robot}

We evaluated our approach in two different platforms (both open and standard), namely robot Justina -- a domestic service robot developed at the Bio-robotics Laboratory in the National Autonomous University of Mexico \cite{Matamoros_2013}, \cite{Savage_2016} and the Toyota Human Support Robot (HSR) -- a service robot developed by Toyota Company \cite{hashimoto:2013} --, and we use the Robot Operating System (ROS) \cite{quigley:2009}. This implementation allowed us to be one of the two teams with highest score in the standard category.

In our implementation, after the objects on the table are detected and recognised, this information is sent through ROS messages to the action planner module. The message contains information about each object, such as: a) related to the point cloud: eigenvalues, eigenvector, size, nearest point and centre point in robot coordinates, b) related to the image: position of top left corner, width and height of the bounding box, c) related to the object: centroid, roll, pitch and yaw manipulation angles, and a priority constant (this variable is set given the grasping ability of a given object with the robot hand -- the priority, in increasing order, is as follows: dish, cutlery, bowl and glass).  

%\textbf{TODO - How are items picked up? Whats decisions do the robot make?}\\

To decide which object to take first, two measures are taken into consideration: the robot-to-object distance and the manipulation priority constant, both measures are combined to decide which object to take first, through manipulation process described below.

\subsection{Object Manipulation} 

\begin{figure}[h]
\begin{center}
\resizebox{0.45\textwidth}{!}{
		$\begin{array}{c}
		\includegraphics[width=90mm]{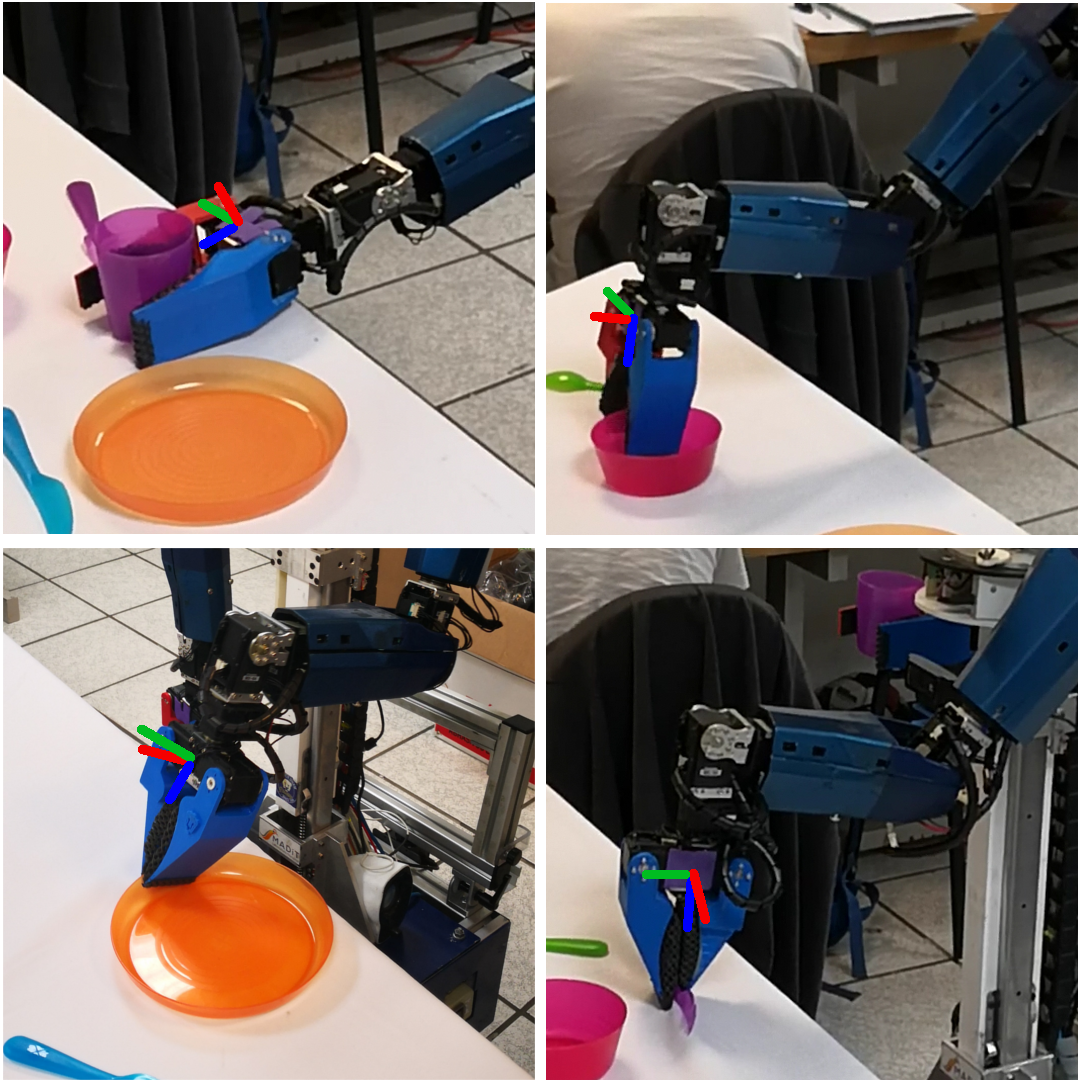} 
		\end{array}$}
	\end{center}
	\caption[]{Gripper orientation and reference system for different objects, top left to bottom right, cup, bowl, plate, and cutlery. The red, green and blue arrows are the manipulator X, Y, and Z axes, respectively.}
    \label{fig:justinaManip}
\end{figure}

The robot end-effector roll, pitch and yaw angles are determined according to the object class to be manipulated, as shown in Figure \ref{fig:justinaManip}. Considering a table plane XY with normal Z, in the case of glasses, 
%it is natural for the robot hand to reach its final position as aligned as possible to the object,
the manipulator moves parallel to the plane XY towards the object's centroid, i.e. its yaw angle is 90 degrees and the other two angles are zero. For plates and bowls, the best way to handle them is by taking them from above in any point on the edge, so the end-effector moves in the Z axis in the direction of a selected point on the object's edge, i.e. its three angles are zero. Finally, to manipulate cutlery, the robot manipulator grasps them from above, where the hand moves in the Z axis centred at the cutlery centroid, i.e. pitch and yaw angles are zero, and the manipulator's roll angle is parallel to the direction of the cutlery's principal component, according to the Equation (\ref{eq1}).

\begin{equation} \label{eq1}
roll=atan2(e_y,e_x)
\end{equation}

where $e_x$ is the X coordinate and $e_y$ is the Y coordinate of the principal eigenvector in robot hand coordinates.
    
The object manipulation process is as follows. First, the robot moves its base such that the arm is located just in front of the object to be manipulated. Then, from the inverse kinematics models, we calculate the arm joint angles to reach the goal position as well as the end-effector orientation, according to the object class to be manipulated, as described above. 

%\textbf{FIX: I didn't understand this paragraph (maybe it's not necessary?):}
%Reynaldo Martell: 
%\textbf{for me it's understandable, but if you think it's not necessary, it's fine}
Different cases are handled in the action planer regarding the end-effector trajectories depending on the object to be manipulated. In the case of glasses and cups, as the robot hand moves on the table plane and trajectories are drawn in a straight line from the robot manipulator to the object centroid, we place the hand at a safe height such as it does not collide with the table but is still able to reach the object. For the other objects, the robot hand moves in the Z axis (normal to the table plane) so the manipulator is placed at a safe distance over a target object and slowly decreased until it reaches the goal position.

\subsection{Visual Feedback}

For conventional objects such as cans, bottles, or small boxes, our robot uses as feedback the torque exerted by the motors on the object to know if the object was taken successfully. However, due to the size and shape of the objects in this test, we use a different feedback strategy based on the visual information to verify whether the object was taken.

%Once the manipulation task has been completed, we verify that the task has been carried out successfully. For this we do the following: 
As a result of the recognition process, we know the object's class, colour, centroid, and bounding box, and we use this information to validate that the object has been taken by finding whether the object is still present on the surface where it should have been removed (for example a plate on the table). We look for the closest objects to the target centroid on the table after the manipulation process and, if an object with the same class and colour is present within a valid distance to the target centroid, we infer that the grasping process failed and start it again.
%; for this, once the manipulation task is finished, we look for if the object in question is still there, how do we determine it? Simple, we look for in the arrangement of objects that we observe in the current world with the vision system an object with the same characteristics (colour and type) as the object we wanted to manipulate, if there is no object with these characteristics we can say that the object was taken, if there is an object with the same characteristics we must now make sure that it is the same object to do so we look if the centroid of the object that we are trying to manipulate is within the bounding box of the objects that have the same characteristics of colour and type at the moment of verify the information, if the centroid is inside the bounding box we can say that it is the same object and the robot will know that the task failed and try to manipulate that object again, if the centroid is not inside the bounding box it means that it is of a different object and therefore the robot will know that it did the task satisfactorily and will go on to perform another task such as taking another object, navigating, etc.

\section{RESULTS IN COMPETENCE}

%\textbf{TODO - Describe arrangement of tableware and cutlery.}

In the PGDC test, robot "Justina" obtained a good performance in the manipulation tasks, as we can see in TABLE \ref{tab:tableJustina} where the robot was able to grasp 3 out of 4 objects that the robot attempted to manipulate, two of them in the first attempt, and the other in the second attempt after visual feedback confirmation was received. Similarly, robot "Takeshi" managed to grasp 2 out of 3 objects that it tried to manipulate, as shown in TABLE \ref{tab:tableTakeshi}. 

\begin{table}[h!]
  \begin{center}
    \caption{Robot Justina's manipulation results.}
    \label{tab:tableJustina}
    \begin{tabular}{c|c|c}
      \textbf{Type} & \textbf{Successful Graspping} & \textbf{Attempts to grasp}\\
      \hline
      glass & yes & 1 \\ 
      dish & yes & 1\\ 
      cutlery & yes & 2\\ 
      cutlery & no & 3\\ 
    \end{tabular}
  \end{center}
\end{table}

\begin{table}[h!]
  \begin{center}
    \caption{Robot Takeshi's manipulation results.}
    \label{tab:tableTakeshi}
    \begin{tabular}{c|c|c}
      \textbf{Type} & \textbf{Successful Graspping} & \textbf{Attempts to grasp}\\
      \hline
      bowl & yes & 1 \\ 
      dish & no & 1\\ 
      dish & yes & 1\\ 
    \end{tabular}
  \end{center}
\end{table}

\begin{figure}[h]
\begin{center}
\resizebox{0.50\textwidth}{!}{
		$\begin{array}{c}
		\includegraphics[width=50mm]{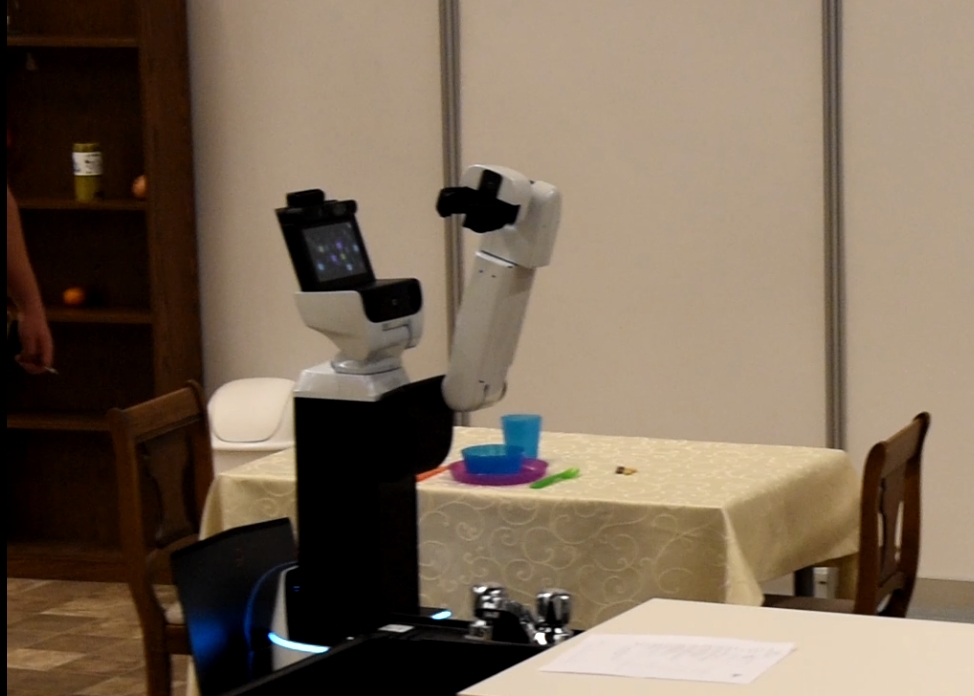} 
		\end{array}$}
	\end{center}
	\caption[]{Robot Takeshi (Toyota HSR) at the RoboCup@Home P\&G Dishwasher Challenge. The robot moves from a known position and aims at cleaning the table.}
	\label{fig:grasping}
\end{figure}

It is important to notice that the manipulator in robot "Takeshi" (Toyota HSR, as in Figure \ref{fig:grasping}) is more precise at the hardware level than the one in robot "Justina", but visual feedback helps to improve the performance in both cases to the point to have similar outcomes, a result that shows the relevance of using intelligent manipulation strategies regardless of the hardware architecture.

\section{CONCLUSIONS}

We presented our approach to the flat and textureless object manipulation problem. We use colour and geometry cues from RGBD sensors, and define different intelligent grasping strategies depending on the object class. Furthermore, we use a visual feedback to determine whether the grasping process has been successful. The described strategies were used in the RoboCup@home league competition under an open and standard platform, having one of the best performances. For future work, we aim at using active feedback models to update in real time the manipulator position with respect to the target object and the grasping success/failure belief.

\addtolength{\textheight}{-12cm}   % This command serves to balance the column lengths

%%%%%%%%%%%%%%%%%%%%%%%%%%%%%%%%%%%%%%%%%%%%%%%%%%%%%%%%%%%%%%%%%%%%%%%%%%%%%%%%

%%%%%%%%%%%%%%%%%%%%%%%%%%%%%%%%%%%%%%%%%%%%%%%%%%%%%%%%%%%%%%%%%%%%%%%%%%%%%%%%

%%%%%%%%%%%%%%%%%%%%%%%%%%%%%%%%%%%%%%%%%%%%%%%%%%%%%%%%%%%%%%%%%%%%%%%%%%%%%%%%
%\section*{APPENDIX}
%TODO - if it's necessary

%\section*{ACKNOWLEDGMENT}
%TODO - if it's necessary

%%%%%%%%%%%%%%%%%%%%%%%%%%%%%%%%%%%%%%%%%%%%%%%%%%%%%%%%%%%%%%%%%%%%%%%%%%%%%%%%

%----------------------------------------------------------------------------------------
%	BIBLIOGRAPHY
%----------------------------------------------------------------------------------------

\bibliographystyle{ieeetr}
\bibliography{all}
%ADD REFERENCES IN THE all.bib FILE IN BIBTEX FORMAT!!!

\end{document}